\documentclass[letterpaper, 10 pt, conference]{ieeeconf}  

\IEEEoverridecommandlockouts                              

\usepackage{hyperref}
\usepackage{graphicx}
\usepackage{amsmath}
\usepackage{cite}

\title{\LARGE \bf
Minimizing Human Assistance: Augmenting a Single Demonstration for Deep Reinforcement Learning}

\author{Abraham George$^{1}$, Alison Bartsch$^{1}$, and Amir Barati Farimani$^{1}$
\thanks{$^{1}$With the Department of Mechanical Engineering,
        Carnegie Mellon University 
        {\tt\small \{aigeorge, abartsch, afariman\} @andrew.cmu.edu}}%
}

\begin{document}

\maketitle
\thispagestyle{empty}
\pagestyle{empty}

\begin{abstract}
The use of human demonstrations in reinforcement learning has proven to significantly improve agent performance. However, any requirement for a human to manually 'teach' the model is somewhat antithetical to the goals of reinforcement learning.  This paper attempts to minimize human involvement in the learning process while retaining the performance advantages by using a single human example collected through a simple-to-use virtual reality simulation to assist with RL training. Our method augments a single demonstration to generate numerous human-like demonstrations that, when combined with Deep Deterministic Policy Gradients and Hindsight Experience Replay (DDPG + HER) significantly improve training time on simple tasks and allows the agent to solve a complex task (block stacking) that DDPG + HER alone cannot solve. The model achieves this significant training advantage using a single human example, requiring less than a minute of human input. Moreover, despite learning from a human example, the agent is not constrained to human-level performance, often learning a policy that is significantly different from the human demonstration.

\end{abstract}


\section{INTRODUCTION}

Reinforcement learning (RL) is widely used to allow robots to autonomously learn how to complete tasks, such as grasping and manipulation \cite{zeng2018, quillen2018, vecerik2019, stulp2012, popov2017, billard2018}, walking \cite{morimoto2004, rudin2022, haarnoja2018, yang2020}, aerial robotic flight \cite{sampedro2018, koch2019}, and many others. However, the training process is often both time and resource intensive, even for relatively simple tasks. A recent survey paper highlighting the challenges in the field of deep reinforcement learning (DRL), \cite{ibarz2021} noted sample efficiency, reliable and stable learning, and side-stepping exploration challenges all as current bottlenecks. As such, finding ways to improve the stability and learning speed of reinforcement learning models is vital to their widespread use. One such way is to provide human examples of the task being completed to help guide the reinforcement learning algorithm \cite{nair2018}. Although human examples have proven to be effective at increasing the training speed and accuracy of DRL algorithms \cite{nair2018, vecerik2017, zuo2017, hester2017}, they are somewhat counter-productive, as one of the major advantages of DRL is that it can complete tasks without requiring an expert to spend time implementing a controller, or generating numerous demonstrations. Additionally, training human experts to use demonstration methods that provide rich information, such as teleoperation, can be time-consuming. However, this challenge can be addressed with virtual reality based teleoperation, because in the virtual space a human can intuitively perform actions without much difficulty, while still generating rich trajectory information.

\begin{figure}[thpb]
      \centering
      \includegraphics[width=\linewidth]{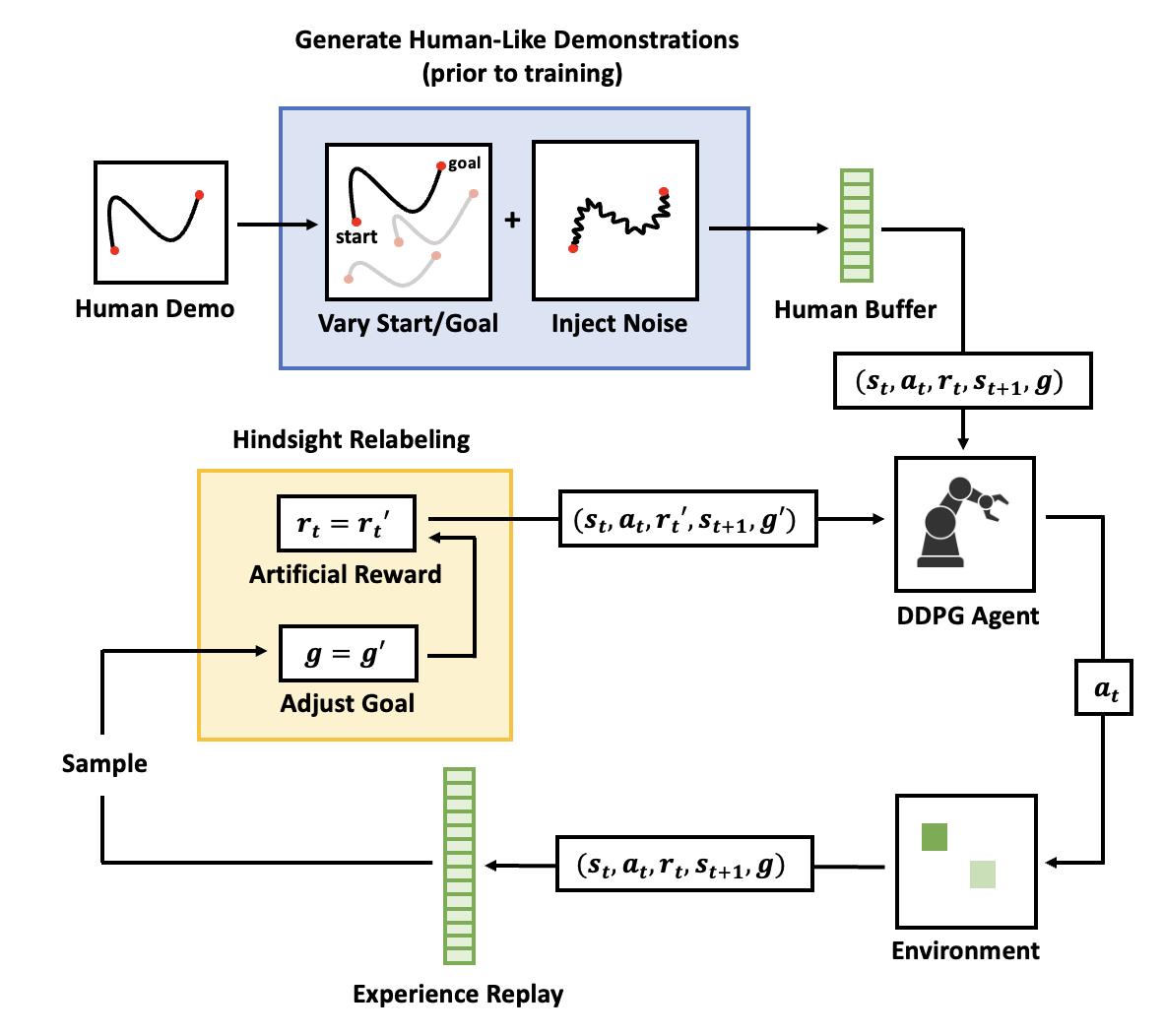}
      \caption{\label{fig:flowchart}A diagram outlining the components of our method. It consists of a standard DDPG + HER architecture that is augmented with a secondary buffer populated with a variety of human-like trajectories, which are generated from a single collected human demonstration.}
      \label{figurelabel}
  \end{figure}

In this paper, we propose a middle ground solution to these challenges: a method that uses a single human example, collected through a simple-to-use virtual reality (VR) simulation, to assist with DRL training. By limiting the human input as much as possible, this paper endeavors to gain significant training advantages while requiring less than a minute of human input. The key aspects of our method are

\begin{itemize}
    \item A demonstration trajectory collection method that is easy and intuitive to use, leveraging VR technology.
    \item A trajectory generation method that augments the single human demonstration by varying the start and goal positions and injecting noise. It is used to populate the demonstration buffer with many varying demonstration trajectories created from a single human example.
    \item A DDPG + HER architecture which is combined with the demonstration buffer to train the agent with no additional reward engineering necessary.
\end{itemize}

A schematic diagram outlining this architecture can be seen in Fig. \ref{fig:flowchart}. The primary technical contribution of this work is the development of an augmentation method that, given a single human demonstration, can generate a diverse set of trajectories to train a DRL agent. Our system converges to successful policies on standard manipulation tasks faster than the baseline of Deep Deterministic Policy Gradients (DDPG) \cite{lillicrap2015} with Hindsight Experience Replay (HER) \cite{andrychowixz2017}, while requiring minimal additional effort to collect a single demonstration. Additionally, we demonstrate that our method can achieve success on more complicated and longer time-horizon tasks, such as block stacking, that standard DDPG + HER is unable to learn. 

\section{Related Works}

\subsubsection{Imitation Learning}
Imitation learning uses demonstrations to determine an agent's actions by having the agent replicate expert examples. This task can be accomplished through machine learning using a supervised learning approach such as classification \cite{ratliff2007}. These forms of behavior cloning have proven effective on complex tasks such as biped locomotion \cite{nakanishi2004}, but they require large data sets and do not function well in environments outside of the range of examples they trained on. Data set Aggregation (DAgger) addresses some of theses issues by augmenting the learned policy with expert demonstrations collected throughout the training process, interconnecting the learned and expert policies \cite{ross2011}. However, DAgger can be difficult to implement because it requires the use of expert examples collected throughout the duration of training.

\subsubsection{Human Demonstrations in Reinforcement Learning}
Human demonstrations have proven a powerful aid in augmenting reinforcement learning. By combining human demonstrations with reinforcement learning, classical control tasks such as the pendulum swing up task can be learned \cite{atkeson1997}. Human demonstrations have been combined with Deep Q-learning to significantly accelerate learning, even with relatively small amounts of data \cite{hester2017}. This Deep Q-learning approach has also been extended to the continuous action space, where demonstrations have been shown to aid DDPG in learning to drive a car in a simulated environment \cite{zuo2017}. Additionally, demonstration learning has proven effective in assisting with robotic sparse reward tasks, such as peg insertion \cite{vecerik2017}. \cite{delacruz2019} found moderate success pre-training with non-expert human demonstrations, first the authors used supervised learning to learn task features from human demonstrations, then combined this pre-trained feature network with asynchronous advantage actor-critic to learn a policy for the Atari domain. Notably, this method still improved over the baseline with low-quality demonstrations. In \cite{taylor2011}, researchers present an algorithm that transfers 20 human demonstrations into a baseline policy that is refined with reinforcement learning, which produces small but statistically significant improvement in performance. They found the best performance when combining the best demonstrations from two teachers, highlighting the need for high quality demonstrations for this technique. However, neither \cite{delacruz2019} nor \cite{taylor2011} significantly improved over the baselines, while both required multiple human demonstrations. A different vein of work focuses on using human preferences and feedback to shape learned policies. \cite{griffith2013} presents a Bayesian approach to maximize the information gain from the human feedback, converting this feedback into a policy, and combining this policy with a RL policy. In \cite{christiano2017}, the goal of the agent is defined through human preferences between trajectories. They demonstrate the method was able to solve a suite of simulated robotics tasks without a reward function. However, each task required an hour of human time.

In this work, our experiments closely match those done by \cite{nair2018}, where a data set of 100 human demonstrations was used to significantly increase the performance of deep reinforcement learning agents on robot manipulation tasks, including pick-and-place and block stacking. The human examples were added to the replay buffer, the loss function was adjusted with a behavioral cloning component to incentivize the agent to match the human demonstrations, and the agent was initialized in demonstration states to assist with the long-horizon sparse reward tasks, in this case block stacking. The primary limitation of \cite{nair2018} is the necessity of collecting 100 human demonstrations for each task. Our work seeks to replicate the successes while only using a single human example and not requiring any modifications to the loss function.

\section{METHODS}

\subsection{Human Example Collection and Augmentation}

\begin{figure}[thpb]
      \centering
      \includegraphics[width=\linewidth]{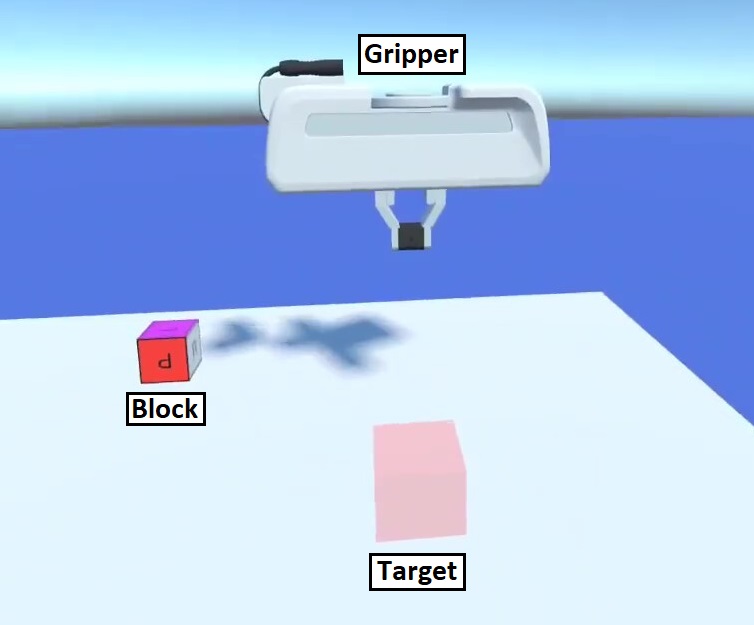}
      \caption{\label{fig:oculus_view}View from the Oculus VR headset of the pick-and-place task. The user moves the gripper with a hand-held controller, and is tasked with moving the block to the target location.}
      \label{figurelabel}
  \end{figure}

A Meta Oculus Quest 2 virtual reality (VR) headset was used to collect human examples of the reinforcement learning (RL) tasks being completed. The VR environment shows the user a first person view of the task they needed to complete. For example, when demonstrating the pick-and-place task, the user is shown the block that needs to be manipulated, along with a highlighted goal region where the block is to be placed. A view of this environment can be found in Fig. \ref{fig:oculus_view}. To manipulate the blocks, the user controls a Franka Emika Panda \cite{franka2021} end effector, which is mapped onto the right hand Oculus controller, and whose gripper is opened and closed with the controller joy-stick. The VR environment records the user's actions when completing the desired task, creating a trajectory file that can then be used to generate a replay buffer of human actions.

Although only a single human example is sampled, a wide range of demonstrations with different start and goal conditions, along with different approaches to solving the task, is ideal for training the reinforcement learning algorithm. To help increase the diversity of our human samples, a method to generate human-like examples from the single recorded human example was developed. First, random object start and object goal locations are chosen. The recorded trajectory is then linearly scaled such that the recorded start and goal locations match the new locations. For the stack task, which has multiple object start and goal locations, the trajectory is split into two sub-trajectories (move block one, then move block two) and each sub-trajectory is scaled independently. The robot then traverses the task environment using the generated trajectory and a proportional controller, and the resultant actions, rewards, and states are recorded. Although this scaling allows for the generation of new trajectories, it does not add much variety. To decrease the trajectory similarity, Ornstein–Uhlenbeck noise \cite{nauta2019} is added to the robot's actions. The augmented trajectories that successfully complete the task are used to populate a secondary replay buffer (referred to as the human buffer) which can then be used to train the reinforcement learning algorithm. The process for generating human-like actions is shown below in equations 1-4.

\begin{equation}
    a = 
    \begin{cases}
    \frac{p_{gen,goal} -  p_{gen,start}}{p_{rec,goal} -  p_{rec,start}},& \text{if } p_{rec,goal} \neq  p_{rec,start}\\
    1,              & \text{otherwise}
    \end{cases}
 \end{equation}
 \begin{equation}
    b = {p_{gen,start}} - a \cdot p_{rec,start}
 \end{equation}
 \begin{equation}
    {p_{gen,i}} = a\cdot{p_{rec,i}} + b
\end{equation}
\begin{equation}
    action_{p,i} = K (p_{gen,i} - p_{measured,i}) + N_{OU}
\end{equation}

Where $p$ is a trajectory coordinate (the process is run for $p = x, y$, and $z$), $rec$ refers to the recorded trajectory, $gen$ refers to the trajectory that is to be generated, $goal$ refers to the goal block location, $start$ refers to the start block location, $i$ is the time step, $K$ is the proportionality constant for the controller, $p_{measured,i}$ is the measured location of the end effector at time step $i$, and $N_{OU}$ is Ornstein–Uhlenbeck noise.

\subsection{Learning Architecture}

Our human aided reinforcement learning method follows the general DDPG + HER DRL structure, except with a secondary replay buffer storing the augmented human demonstrations. During training, 25\% of the transitions are sampled from this secondary human buffer, while the remaining 75\% are sampled from the standard experience replay buffer. This method, which is similar to the replay buffer spiking introduced by Lipton et. al, \cite{lipton2016} exposes the RL agent to the human example without tying it to the human developed policy. Although previous work has shown that directly linking human demonstrations to the training of the agent through methods such as using behavior cloning loss can be effective, \cite{nair2018} we fear that the lack of diversity in our human samples may cause this type of approach to lead to overfitting.

Additionally, a demonstration based pre-play method was implemented for use in tasks with a long time horizon and a sparse reward, in this case block stacking. This type of task is particularly challenging because the agent needs to learn a more complex sequence of actions before it receives a reward. The pre-play method addresses this issue by initializing the training environment using actions based on the human demonstration. In 50\% of the simulation episodes, the initial conditions of the environment are used to generate a human-like trajectory, which the robot follows for the first $n$ time steps, where $n$ is randomly generated from a uniform distribution between $0$ and the total number of simulation steps. After the $n$ steps, the robot's actions are determined by the DDPG agent. This approach provides the agent with simplified versions of the task by having the demonstration based pre-play partially solve the task. By exposing the agent to the reward early in the training process, we hope this method will function similarly to curriculum learning and help improve the agent's performance. Additionally, since a method to generate human-like examples for a given start and goal location has been developed, the human-like trajectory used to determine the action policy before step $n$ can be tailored to the specific environment conditions, allowing this method to be used when training on hardware.

\section{EXPERIMENTAL EVALUATION}

\subsection{Simulation Environment}

\begin{figure}[thpb]
      \centering
      \includegraphics[width=\linewidth]{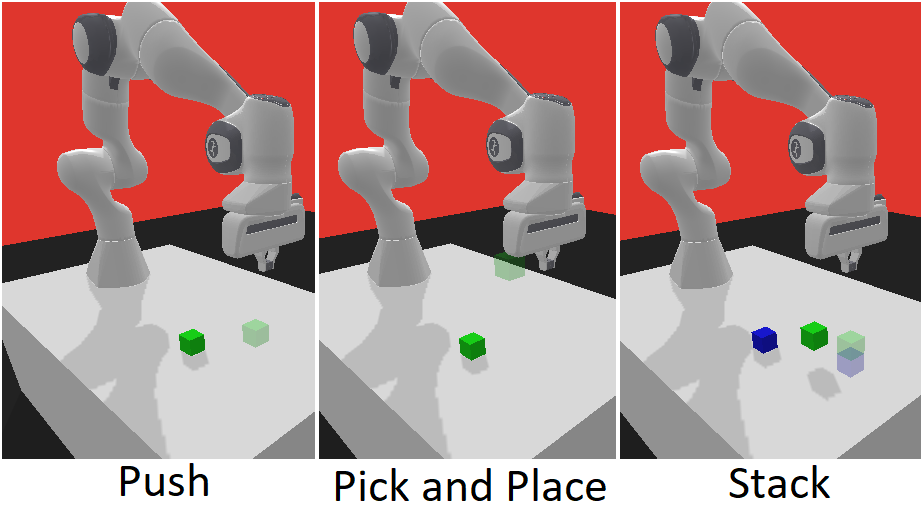}
      \caption{\label{fig:gym_env}Panda-Gym environments used for RL model evaluation. The block(s) to be stacked are shown as opaque cubes, while the goal locations are shown as transparent cubes.}
      \label{figurelabel}
  \end{figure}

The RL model was evaluated on three Panda-Gym tasks created by Gallouedec et al.: push, pick-and-place, and stack, as shown in Fig. \ref{fig:gym_env} \cite{gallouedec2021}. The panda-gym tasks, which run on the PyBullet physics engine, feature a 7-degree-of-freedom Franka Emika Panda robot arm which is used to manipulate a 4 cm cube block. During the RL experiments, the orientation of the gripper is locked, making the end effector a 4-degree-of-freedom system (position and gripper width). A sparse reward was used during training, returning a reward of $-1$ when the task is not complete and $0$ when the task is complete. Task completion is determined by taking the distance from the current cube location (opaque cube in Fig. \ref{fig:gym_env}) to the goal cube location (transparent cube in Fig. \ref{fig:gym_env}), and checking if that distance is less than the cutoff distance (5 cm in the push and pick-and-place tasks and 4 cm for both cubes in the stack task). Two modifications were made to Gallouedec et al.'s Panda-Gym tasks. In the push task, the fingers were unlocked so that they could be used to manipulate the block, and in the stack task, the completion criteria was changed from both cubes having less than 5cm of error to both cubes having less than 4cm of error to remove the possibility of false positive rewards. To evaluate our method's performance, all experiments were conducted across five or more random seeds.

From the Panda-Gym simulation, the DRL agent observes the following information: the end effector position and velocity, the gripper width,  and the block(s) position, rotation, and velocity (both translational and rotational). Given this state information and the goal, the DDPG agent selects an action (the output of the actor network). This action has 4 parameters, the x,y,z displacement to apply to the end effector and the displacement to apply to the width between the gripper fingers. When training the DDPG agent, each epoch included 1280 simulations (50 time-steps each for push and pick-and-place, and 100 time-steps for stack) and 3200 network updates (batch size of 256). 

\subsection{Results}
\subsubsection{Effect of the Human Buffer}

\begin{figure}[thpb]
      \centering
      \includegraphics[width=\linewidth]{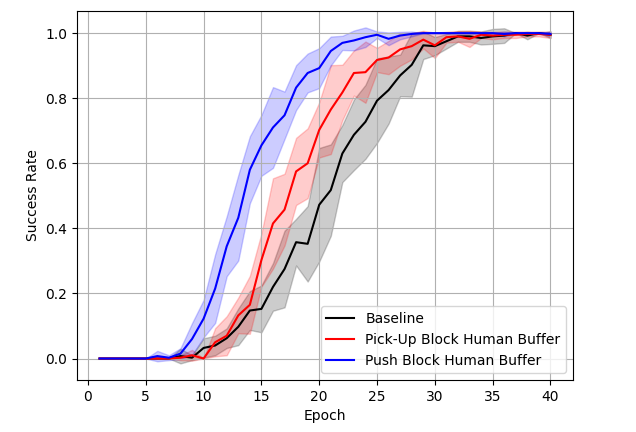}
      \caption{\label{fig:push_human_v_base}Success rate on the push task for baseline DDPG + HER, DDPG + HER with the human buffer created from a pick-up demonstration, and DDPG + HER with the human buffer created from a block pushing demonstration. The shaded region shows one standard deviation from the mean. The introduction of the human buffer causes the model to train faster, even when the demonstration is significantly different from the converged policy.}
      \label{figurelabel}
  \end{figure}

\begin{figure}[thpb]
      \centering
      \includegraphics[width=\linewidth]{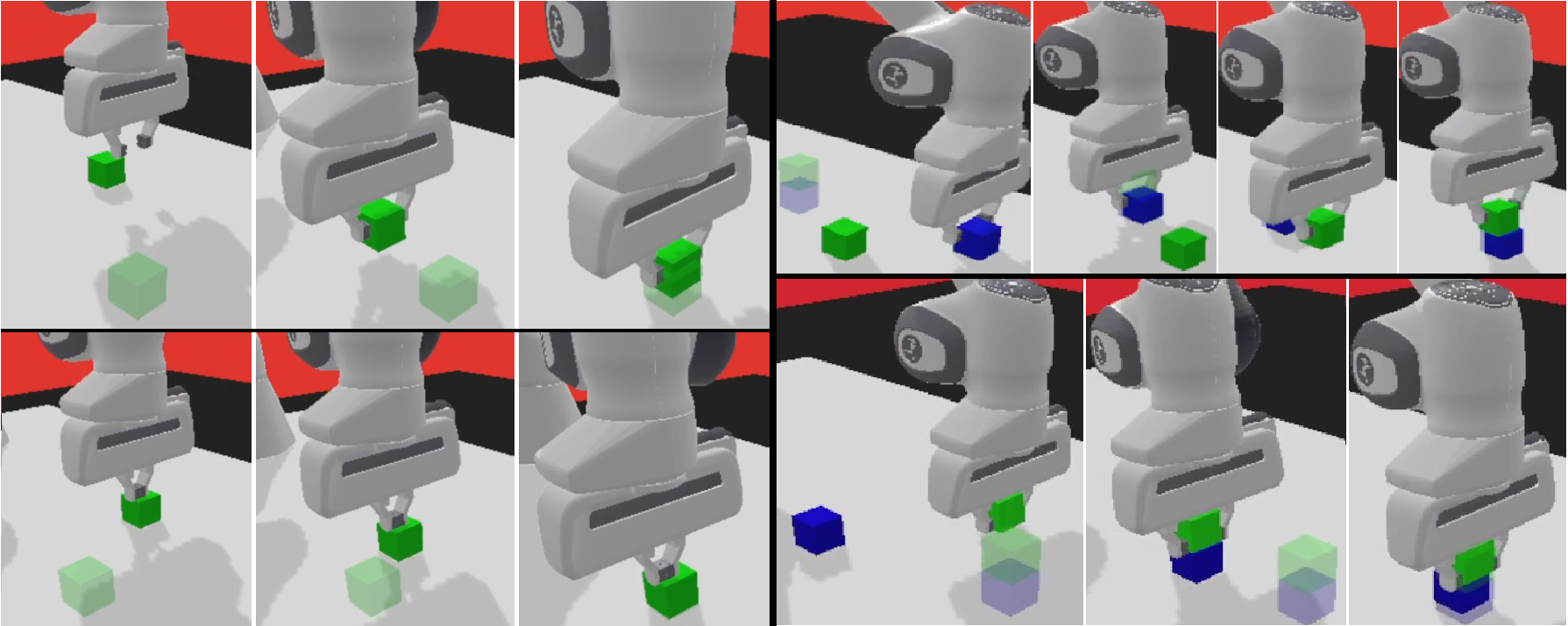}
      \caption{\label{fig:combo_policy}Push and Stack agent policies. Left - Push task: The human demonstrations (top three images) feature picking up and then placing the bock, while the policy the agent converges upon (bottom three images) is to drag the block to the goal using the gripper. Right - Stack Task: The human example (top four images) first picks and places the bottom block, then picks and places the top block. The trained agent (bottom three images) picks up the top block, sets it onto the bottom block, then drags the stack to the goal location.}
      \label{figurelabel}
  \end{figure}

To examine the effectiveness of the human-aided model, we compared the success rate over epoch of the baseline DDPG + HER with our human-aided DDPG + HER implementation for the push and the pick-and-place tasks. The results of these tests can be seen in Fig. \ref{fig:push_human_v_base} and \ref{fig:ablation}. Our experiments show that the addition of the human buffer led to a significant increase in learning speed, especially in the more complex pick-and-place task. We found that regardless of whether the human buffer was used, the agent learned the same policy. For example, in the push task both the baseline DDPG + HER agent and the DDPG + HER agent with a human buffer learned to close the gripper, place it on top of the block, and drag it to the goal location. The agent with a human buffer learned the same policy even though the provided human example used a significantly different approach: grasp the block using the gripper, pick it up, and then place it down in the goal location (see Fig. \ref{fig:combo_policy}). With this disparity in learning technique in mind, we repeated the pick-and-place experiment, but replaced the original human example with an example where the human drags the block using a closed gripper (the learned policy). When shown this demonstration, the DDPG + HER with human buffer agent significantly increased its performance (see Fig. \ref{fig:push_human_v_base}). From these experiments, we draw the conclusion that the human buffer provides a significant assistance to DDPG + HER, even when the human policy is non-ideal.

\subsubsection{Ablation Study}

\begin{figure}[thpb]
      \centering
      \includegraphics[width=\linewidth]{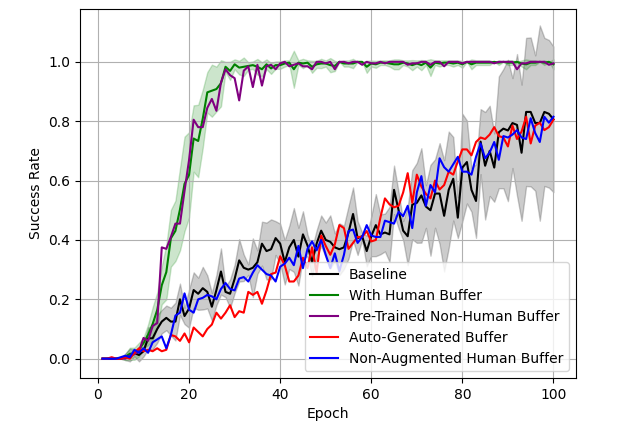}
      \caption{\label{fig:ablation}Success rate on pick-and-place task vs. epoch for baseline DDPG + HER, DDPG + HER with the human buffer, DDPG + HER with a non-human buffer (created by trained agent), DDPG + HER with a generated buffer (created during training), and DDPG + HER with a human buffer without augmentation (only the single recorded trajectory). The shaded region shows one standard deviation from the mean, and was not displayed for all experiments to preserve figure readability. The results show that the augmented human demonstration method significantly outperforms DDPG + HER. Additionally, having an expert demonstration and augmenting that demonstration are vital to our method, but the expert demonstration does not have to be human generated. }
      \label{figurelabel}
\end{figure}

To examine the impact of the human-generated buffer on the results, we performed an ablation study to analyze the key aspects of our approach: the example collection method, the augmentation step generating human-like samples, and the necessity of an initial demonstration. First, to test if the human buffer aided the agent’s learning because it consisted of transitions that were collected from a human demonstration, as opposed to those created by the RL agent, the pick-and-place experiment was re-run using a trained DDPG + HER agent to create the ‘human’ example. As shown in Fig. \ref{fig:ablation}, this method performed similarly to its human counterpart, showing that the example trajectory does not need to be a human demonstration. Next, to test if the augmentation step of generating human-like samples was necessary, the pick-and-place task was run with a human buffer that only contained the original recorded trajectory. This method performed about the same as baseline DDPG + HER, showing that the augmentation step in our method is vital to the agent's performance.

Finally, to test if the increase in performance was due to the generation of new trajectories (augmentation step) rather than the quality of the original successful example, we replaced the human buffer with an auto-generated previous successes buffer that used the agent’s successful runs and our augmentation algorithm to generate example trajectories during training. The concept of leveraging the agent's successes is similar to that of \cite{luo2020}, except in this case these successful trajectories are augmented using our generation method. This approach had a similar success rate to baseline DDPG + HER, suggesting that the expert demonstration itself is also a vital part of the method's success.

\subsubsection{Human-Aided Block Stacking}

\begin{figure}[thpb]
      \centering
      \includegraphics[width=\linewidth]{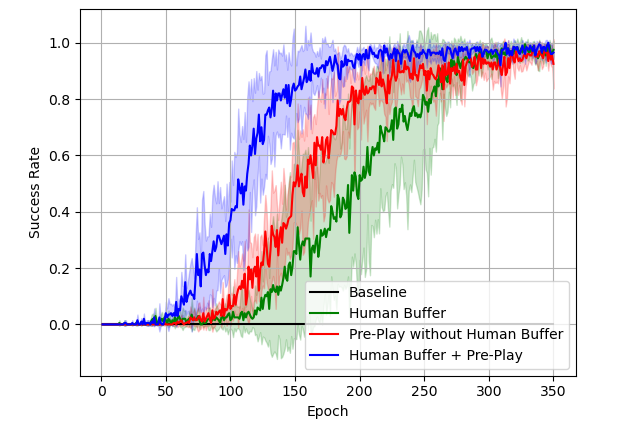}
      \caption{\label{fig:stack_human_v_base}Success rate on the stack task for baseline DDPG + HER, DDPG + HER with the human buffer, DDPG + HER without the human buffer but with demonstration based pre-play, and DDPG + HER with the human buffer and demonstration based pre-play. The shaded region shows one standard deviation from the mean. The results show that both the human buffer and the demonstration based pre-play improve the agent's performance.}
      \label{figurelabel}
  \end{figure}

After analyzing the impact of the single-example human buffer on the push and pick-and-place tasks, we experimented on the stack task. This task, which requires the robot to stack two blocks on top of each other, is more complex than the previous two tasks. Our results can be seen in Fig. \ref{fig:stack_human_v_base}. In this trial, the baseline DDPG + HER approach failed, with a constant 0\% success rate. However, when given a human buffer the agent was successful, converging to above a 90\% success rate in 270 epochs. We also tested DDPG + HER with the demonstration based pre-play method, which managed to reach 90\% success in 230 epochs. By combining the pre-play method with the human buffer, we were able to further increase the agent's performance, reaching  over a 90\% average success rate in only 170 epochs.

As with the push task, the agent learned a policy for completing the stack task that was significantly different from the human example. The human example solves the task by first placing the bottom block in place, then retrieving the top block and stacking it onto the bottom block. However, the RL agent learned to first grab the top block, place the top block onto the bottom block, then use the still grasped top block to drag the stack to the goal location (shown in Fig. \ref{fig:combo_policy}). This discrepancy shows that the demonstration based pre-play approach assists training without restricting the agent to the demonstration policy.

\subsubsection{Effect of HER with Human Buffer}
\begin{figure}[thpb]
      \centering
      
      \includegraphics[width=\linewidth]{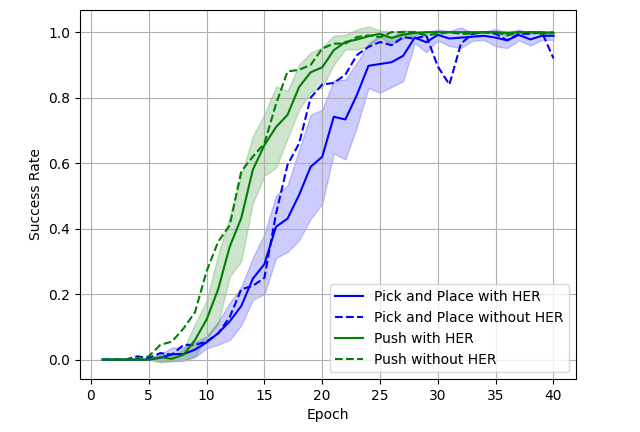}
      \caption{\label{fig:push_pick_her}Effect of hindsight experience replay (HER) on the push and pick-and-place tasks. The shaded region shows one standard deviation from the mean. HER does not seem to have a significant effect on the agent's learning.}
      \label{figurelabel}
  \end{figure}

\begin{figure}[thpb]
      \centering
      \includegraphics[width=\linewidth]{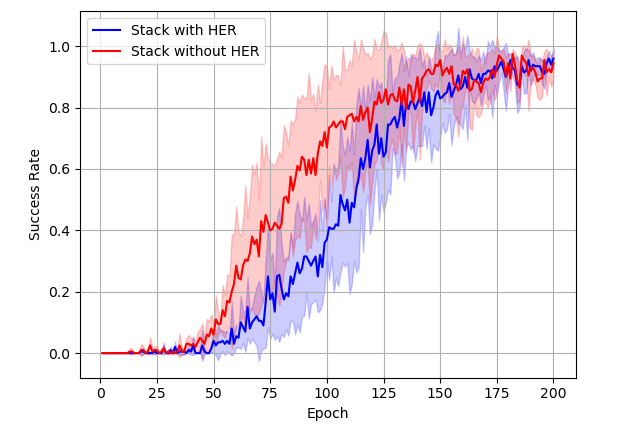}
      \caption{\label{fig:stack_her}Effect of hindsight experience replay (HER) on the stack task. The shaded region shows one standard deviation from the mean. When HER is removed, the agent performs significantly better.}
      \label{figurelabel}
  \end{figure}

Hindsight Experience Replay (HER) is a commonly used technique that assists sparse reward reinforcement learning by creating artificially successful experiences to train on \cite{andrychowixz2017}. These memories are vital because they present the agent with some examples of positive rewards, which in long horizon tasks may be difficult for the agent to reach organically \cite{andrychowixz2017}. However, this role is also fulfilled by both the generated human buffer and the demonstration based pre-play implementation. To determine if the agent still needs HER when using the human buffer, we used DDPG with the human buffer (without HER) to solve the push and pick-and-place tasks, and DDPG with the human buffer and demonstration based pre-play (without HER) to solve the stack tasks (Fig. \ref{fig:push_pick_her} and \ref{fig:stack_her}). Comparing these results to their HER counterparts, we can see that removing HER appears to have slight positive effect on the pick-and-place and push tasks and significantly improves the performance of the stack task. The disparity between our method with and without HER appears to be correlated with the task complexity.

\subsection{Hardware Validation} 
\begin{figure}[thpb]
  \centering
  \includegraphics[width=\linewidth]{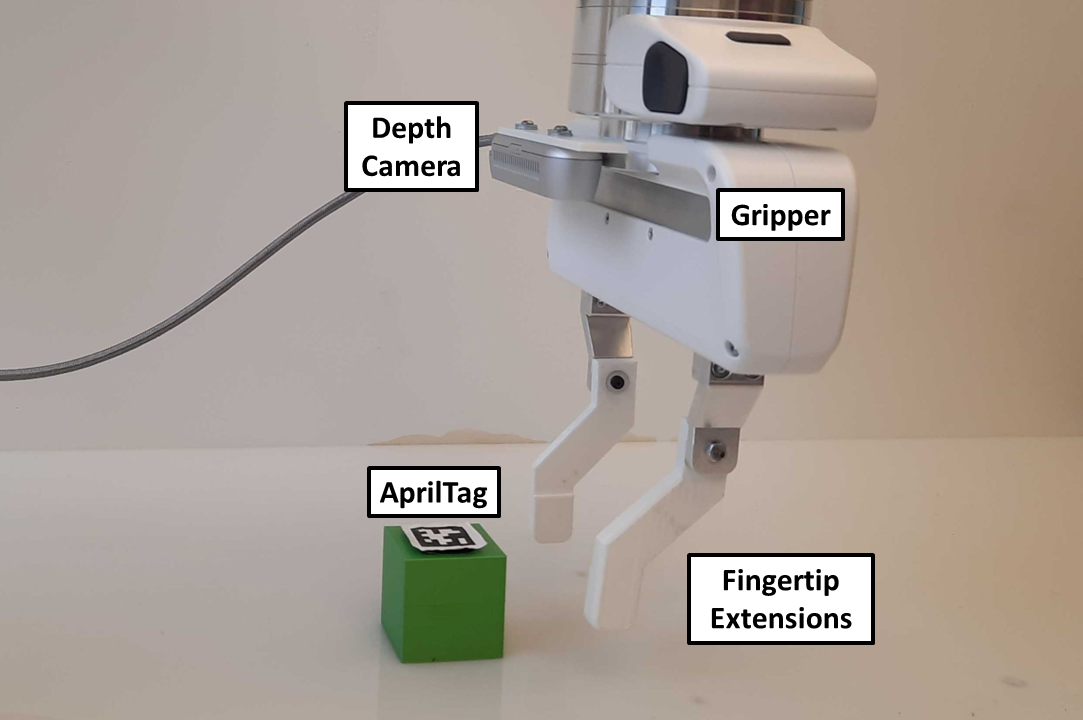}
  \caption{\label{fig:hardware}Hardware validation setup. A Franka Emika Panda arm was used to execute the trained policies. A wrist-mounted RealSense camera was used to track the position of the block(s) using AprilTags. 3D printed finger-tip extensions were used to keep the fingers and blocks centered in the camera frame.}
  \label{figurelabel}
\end{figure}

In order to test the viability of our trained policies for hardware control, we implemented push, pick-and-place, and stack policies, which we had trained in simulation, on a Franka-Emika Panda arm without hardware fine-tuning. A wrist-mounted Intel RealSense D415 RGB-D depth camera was used to monitor the position of the blocks, which were labeled using AprilTags. An image of the hardware setup can be found in Fig. \ref{fig:hardware}. For each task, we tested three separate trained policies and averaged the results. The pick-and-place task performed the best, with an 87\% success rate, while the push task fared worse, with only a 53\% success rate. This discrepancy is likely due to the push task relying on more complicated contact dynamics (sliding the block along the table), which are significantly different between simulation and hardware. Finally, the stack task's learned policy failed on hardware, achieving a 0\% success rate, due to the learned policy's reliance on sliding the stack of blocks along the table, as described in Fig \ref{fig:combo_policy}. When executing on hardware, the agent is able to stack the blocks, but when it tries to slide the stack to the goal location, the stack doesn't slide, instead either toppling or remaining stationary. Overall, these results are promising, especially considering they were achieved with zero-shot sim-to-real transfer. Additionally, we think that the push and stack tasks' performances could be significantly improved with the application of sim-to-real techniques such as hardware tuning and domain randomization during training \cite{pmlr-v100-mehta20a}. 

 \quad  
 
\section{CONCLUSIONS}
Our results show that a single human example can be used to significantly improve DDPG + HER, with the addition of a generated human buffer decreasing the training time (time to 80\% accuracy) of the pick-and-place task to under one fourth of plain DDPG + HER. Additionally, the use of a demonstration based pre-play approach along with the human buffer enabled the block stacking task to be solved. Moreover, because the human example was collected by directly mapping a user’s hand motions in VR, the single demonstration can be completed quickly and without much human effort. Furthermore, our method has been shown to work when the human demonstration is non-ideal, without restricting the agent from finding a more optimal policy. These two aspects combine to allow a non-expert with minimum training to provide all of the necessary human input with a single short demonstration, allowing for a significant increase in model performance from under a minute of human input.

Our augmentation method is in theory  algorithm-agnostic, as it can be paired with any DRL algorithm that employs a replay buffer, such as Soft Actor Critic \cite{sachaarnoja2018} and Twin Delayed DDPG \cite{fujimoto2018}. Thus, moving forward we hope to evaluate the impact of the human-like buffer when combined with these other DRL algorithms to determine if the improvement from the augmentation method holds regardless of algorithm.

Additionally, we would like to improve our model's generalizability, specifically by developing alternative, more robust methods for generating the human-like trajectories. Although the current generation method works well, it requires the task to be composed of sub-tasks that can be linearly interpolated. By developing a more general method for human-like trajectory generation, we could significantly increase the range of tasks to which the single human demonstration augmentation method can be applied.

\addtolength{\textheight}{-2cm}   


\bibliographystyle{IEEEtran}
\bibliography{ref}

\end{document}